\documentclass[letterpaper, 10 pt, conference]{ieeeconf}  % Comment this line out if you need a4paper

\IEEEoverridecommandlockouts                              % This command is only needed if 
\usepackage{amsmath} % assumes amsmath package installed
\usepackage{amssymb}  % assumes amsmath package installed
\usepackage{graphicx}
\usepackage{color}
\usepackage{soul}
\usepackage{caption}
\usepackage{subcaption}
\usepackage{comment}
\usepackage{url}
\usepackage{hyperref}
\usepackage{copyright}

\title{\LARGE \bf
Generating Causal Explanations of Vehicular Agent Behavioural Interactions with Learnt Reward Profiles
}

\author{Rhys P. M. Howard$^{1}$, Nick Hawes$^{1}$, and Lars Kunze$^{1,\,2}$% <-this % stops a space
\thanks{This work was supported by the EPSRC project RAILS (grant reference: EP/W011344/1).}% <-this % stops a space
\thanks{$^{1}$Rhys P. M. Howard, Nick Hawes, and Lars Kunze are with Oxford Robotics Institute, Dept. of Eng. Sci.,
        University of Oxford, 17 Parks Road, Oxford, OX1 3PJ, UK
        }%
\thanks{$^{2}$Lars Kunze is with the Bristol Robotics Laboratory, T-Block, UWE Bristol, Bristol, BS16 1QY, UK
        }%
        \thanks{Correspondence email: {\tt\small rhyshoward@live.com}}
}

\begin{document}

\maketitle
\thispagestyle{empty}
\pagestyle{empty}

%%%%%%%%%%%%%%%%%%%%%%%%%%%%%%%%%%%%%%%%%%%%%%%%%%%%%%%%%%%%%%%%%%%%%%%%%%%%%%%%
\begin{abstract}

Transparency and explainability are important features that responsible autonomous vehicles should possess, particularly when interacting with humans, and causal reasoning offers a strong basis to provide these qualities. However, even if one assumes agents act to maximise some concept of reward, it is difficult to make accurate causal inferences of agent planning without capturing what is of importance to the agent. Thus our work aims to learn a weighting of reward metrics for agents such that explanations for agent interactions can be causally inferred. We validate our approach quantitatively and qualitatively across three real-world driving datasets, demonstrating a functional improvement over previous methods and competitive performance across evaluation metrics.
\vspace{-5mm}
\end{abstract}

\copyrightnotice

%===============================================================================

\section{Introduction} \label{sec:introduction}

\begin{figure}[t]
    \centering
    \begin{subfigure}{\linewidth}
        \centering
        \includegraphics[width=\linewidth]{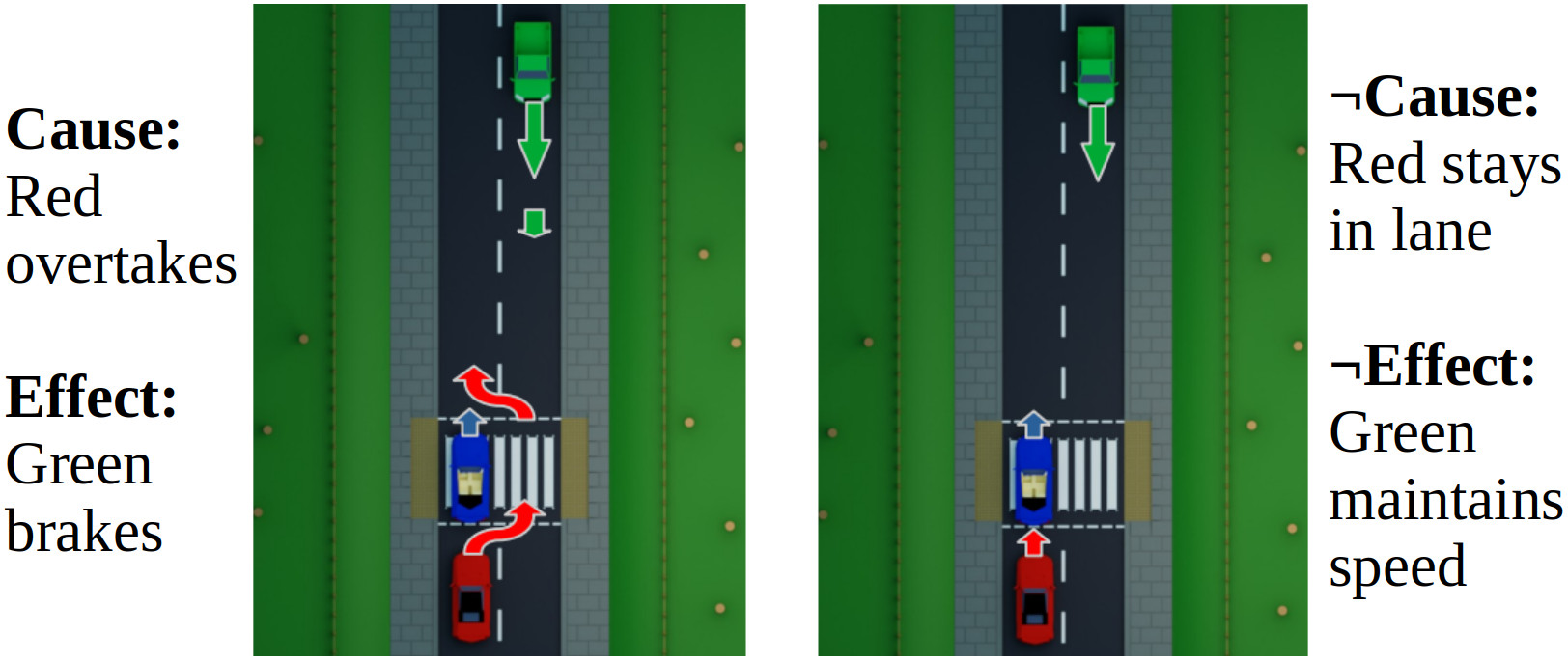}
        \caption{\footnotesize Example of twin-world inference. We can see that toggling whether red overtakes alters green's course of action, thus indicating a causal link. One can also incorporate motivational information from the reward profile. From here it is trivial to generate a textual explanation such as: "Red overtaking caused green to slow down, as green wishes to prioritise safety".}
        \label{fig:example_scenario}
    \end{subfigure}
    \begin{subfigure}{\linewidth}
        \centering
        \includegraphics[width=\linewidth]{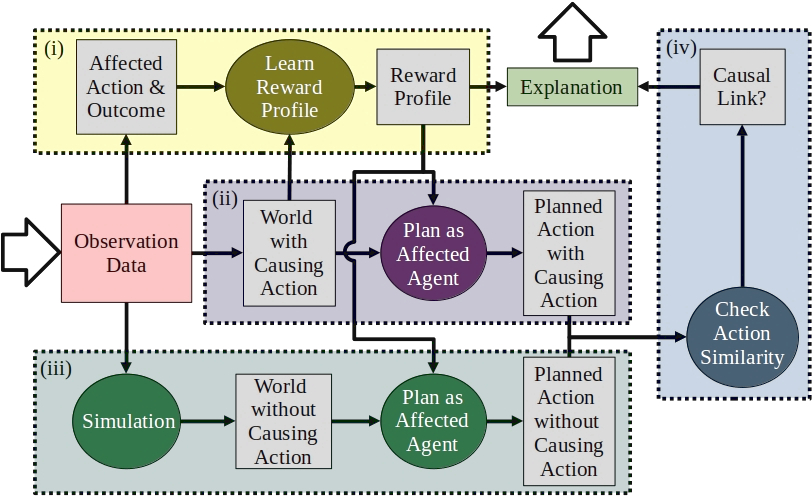}
        \caption{\footnotesize Depiction of our causal explanation generation method. Step (i) learns a reward profile for a primary agent (see Sec. \ref{sec:learning_reward_parameter_profiles}). Step (ii) plans for the primary agent under the observed world state. Meanwhile step (iii) simulates a world in which a secondary agent did not take a particular action, and plans under the resulting world state. Step (iv) compares these these two plans to determine if the secondary agent's action had a causal effect upon the primary agent's behaviour (see Sec. \ref{sec:causal_behaviour_explanation}). The reward profiles and causal link combine to provide a causal explanation of a behavioural interaction.}\vspace{1mm}
        \label{fig:learning_reward_flowchart}
    \end{subfigure}\vspace{-1mm}
    \caption{\footnotesize Illustrations of the proposed method to generate causal explanations for vehicular agent behavioural interactions with learnt reward profiles.}
    \label{fig:twin_world}
\end{figure}

% Motivation
Autonomous systems are becoming increasingly prevalent in our day-to-day lives. Hence we ought to understand cause and effect in relation to their behaviour and the behaviour of others. Autonomous vehicles (AVs) make for a particularly motivating case, as they have a substantial amount of investment currently, yet can also pose a significant risk to human life. This in many ways reflects the early airline industry, and there is work suggesting autonomous systems record their data --- similar to aircraft --- for post-hoc analysis \cite{winfield2022ethical}. Understanding cause and effect here is critical, as many ideas of culpability are tied to causality, not to mention that understanding vehicular agent  behavioural interactions can lead to incremental improvements in the safety of relevant technologies. In order to be able to reason about such interactions effectively one must be able to approximate the instantaneous motivations of agents when making decisions.
%However, in order to be able to reason about autonomous agents as intentional systems \cite{dennett1971intentional} we need to learn what the desires of an agent are at the time of making a decision in order to provide good explanations through causal reasoning \cite{pearl2009causality}.

% What
We build upon our previous work \cite{howard2023simulation} that applies a causally-inspired combination of game theory and theory of mind to discover links between the actions of vehicular agents. Here we utilise a structural causal model (SCM) \cite{pearl2009causality,bareinboim2015bandits} architecture we developed as a case-study in \cite{howard2024extending}. To this we apply contrastive twin-world counterfactual inference (see Fig. \ref{fig:example_scenario}) in order to infer causal links, following the steps depicted in Fig. \ref{fig:learning_reward_flowchart}. Additionally, we now learn and utilise reward profiles to capture agent motivations at the time of decision making, drawing inspiration from inverse reinforcement learning (IRL) \cite{arora2021survey}. Combined with the reward profiles, this produces post-hoc explanations that are more expressive than those produced in our previous work \cite{howard2023simulation}.
%In this work, we present a synthesis of linear regression based approximation of reward metric weightings with twin-world counterfactual inference based explanation generation. Specifically, we utilise reward profile learning in a post-hoc fashion to determine the factors of importance to drivers on the road when making impactful decisions. Doing so allows us to simulate changes in planning resulting from intervening upon the behaviour of other agents, thus establishing the presence of causal behavioural interactions.

% Why
Having given an overview of our motivation and methodology, we state the following as contributions of this work:
\begin{itemize}
    \item Novel integration of one-shot IRL with twin-world counterfactual inference to discover causal links describing behavioural interactions between agents.
    \item Quantitative assessment of the methodology against existing methodologies on the highD \cite{krajewski2018highd} dataset, demonstrating a notable improvement over the previous reward-based approach.
    \item Qualitative assessment of the methodology on the highD \cite{krajewski2018highd}, inD \cite{bock2020ind}, and exiD \cite{moers2022exid} datasets. This illustrates the enhanced expressiveness over the previous approach.
\end{itemize}
%To test the efficacy of our proposed methodology, we run a series of quantitative experiments on the highD dataset \cite{krajewski2018highd}. Here we demonstrate our approach is competitive against the best performing baselines considered, and represents a substantial improvement over the next-best reward-based method. Furthermore we utilise the proposed approach in several qualitative experiments across the highD, inD, and exiD datasets \cite{krajewski2018highd,bock2020ind,moers2022exid} in order to demonstrate the functional improvements of the method over previously proposed techniques. \hl{REWORK / EXTEND THE INTRODUCTION}

%===============================================================================

\section{Related Work}
\subsection{Causal Reasoning in XAI for Autonomous Agents}
Causality has been identified as an important component in the responsible development of systems \cite{mainzer2010causality, kacianka2019extending,llorca2023liability} and recently particular attention has been applied to its use in explainable artificial intelligence (XAI) and robotics \cite{gunning2019darpa,gadd2020sense,hellstrom2021relevance,franklin2022causal}. Of relevance to this work are the sub-fields of algorithmic recourse \cite{karimi2022survey} and temporal causal discovery \cite{assaad2022survey}, both of which offer some level of explanation generation. However, both of these sub-fields have predominantly focused on big-data domains (e.g. medicine, economics, sociology) and work applying causal discovery techniques to real-world vehicle data has shown the unsuitability of many methods for use in the robotics domain \cite{howard2023evaluating}.

Despite this, some works integrate causal reasoning into autonomous embodied agents. Causal models have been applied in offline planning to avoid confounders \cite{cannizzaro2023cardespot}, in online planning to augment reinforcement learning (RL) \cite{bareinboim2015bandits,gasse2022causal,he2023causal}, and to predict failures and take corrective action \cite{diehl2023causal}. In in these cases predefined causal models have been used to bolster existing methodologies, rather than the reverse. There have been works that actively learn causal models as part of their pipeline \cite{yu2023explainable,lin2024safetyaware}, however these methods have relied upon black-box architectures that lack transparency and require large amounts of data.

Of greater relevance here is work concerning the generation of causal explanations, particularly in the context of agent interactions. %Diehl and Ramirez-Amaro \cite{diehl2022why} and Wich \emph{et al.} \cite{wich2022empirical} attempt to explain the behaviours or outcomes associated with single agents, with
\cite{diehl2022why} aims to explain failures for a single robot, while \cite{wich2022empirical} looks to explain the variables that dictate human hand preference for manipulation. 
\cite{castri2022causal} and \cite{tang2022grounded} seek to capture continuous causal relationships within / between agents.
%with Tang \emph{et al.} \cite{tang2022grounded} specifically using this to explain interactions, albeit relying upon a graphical neural network to do so. 
Lastly, \cite{triantafyllou2023towards} evaluates the causal impact of actions, but only for two-player game outcomes.

The works of \cite{maier2023causal} and \cite{gyevnar2024causal} are closest to our own, and we build upon our previous work \cite{howard2023simulation}. \cite{maier2023causal} considers a similar two-vehicle convoy style scenario to this work for the purposes of advanced driver assistance system testing. However, this work relies upon a scenario-specific high level causal model, and depends upon the CARLA simulator \cite{dosovitskiy2017carla} to learn the model's structural equations.
\cite{gyevnar2024causal} offers promising results in offering natural language explanations for action selection from an egocentric perspective. They represent reward and other agents as separate sources of causation and test for causation using the counterfactual effect size model \cite{quillien2023counterfactuals}, differing from our monolithic reward representation and twin-world approach. %Meanwhile we explicitly capture the causal effects of other agents through the reward mechanism of the agent under consideration. Despite this, the egocentric nature and focus upon natural language make quantitative comparison difficult within the context of this work.
Our previous work \cite{howard2023simulation} aims to discover causal links between vehicular agent actions. It uses a game-theoretic comparison of different simulated outcomes without modelling the planning process of agents. However, this is not ultimately grounded in a causal model, nor does it present a mechanism to capture agent motivations. Still, the similarity of the work warrants its use as a baseline in Sec. \ref{subsec:quantitative}.

\subsection{Inverse Reinforcement Learning}
This work bears some resemblance to works in the field of IRL \cite{arora2021survey}, despite not being its primary focus. Here we utilise linear regression to learn a set of weights to be applied to a feature vector that reflect agent motivations.% --- or a ``reward profile" as we refer to it. %Thus rather than a novel contribution to IRL, this work is instead aiming to utilise established IRL methodology in synthesis with causal reasoning to more effectively generate explanations for agent behavioural interactions.

A way in which this work departs from typical IRL is our reward learning is carried out for a single instant in time.
%Note that in contrast to some methods which describe themselves as one-shot \cite{guha2024one} we are not considering a single trajectory, but a single state-action pair. The reward profile intentionally only captures the agent's desires at a given point in time. 
We aim to capture motivations producing dangerous or irrational behaviour that may reflect momentary lapses in judgment. This follows previous work in IRL \cite{kuderer2015learning,arora2021survey} proposing a reward function varying across agents / situations.

It is important to note, that there have indeed been works within causality that do relate to IRL \cite{dehaan2019causal,ruan2023causal}. However, most of this works aim to carry out imitation learning as a means of better performing a task while accounting for confounders, rather than as a means to generate explanations.

\section{Background}

\subsection{Structural Causal Models} \label{subsec:scm}
%In this work we integrate causality into an autonomous system by defining it in terms of SCMs. 
A SCM $\mathcal{M} = (\mathcal{U}, \mathcal{V}, \mathcal{F}, P(\mathcal{U}))$ is defined by exogenous variables $\mathcal{U}$, endogenous variables $\mathcal{V}$, structural equations $\mathcal{F}$, and a probability distribution $P(\mathcal{U})$ for $\mathcal{U}$ \cite{pearl2009causality,bareinboim2015bandits}. Exogenous variables capture unmodelled external factors, while endogenous variables are derived via structural equations, defined for a variable $V$ in terms of its parents $Pa(V)$. %For any given endogenous variable $V \in \mathcal{V}$, its parents $Pa(V) \subset (\mathcal{V} \cup \mathcal{U})$ are comprised of any number of endogenous variables and at most one exogenous variable.

%\subsection{Temporal Extension of SCMs}
By default SCMs do not explicitly capture temporality. The most simple method to integrate such information within the SCM framework is to roll-out the variables across a series of time steps. %Thus for a variable $V$ before roll-out there would be a new set of variables $V_t$ representing the values of $V$ for time steps $t \in \mathbb{N}$.
We assume the modelled system is Markovian with time lags of only a single time step, thus the temporal nature of the SCMs is simple here. We refer the reader to \cite{peters2017elements} for a more comprehensive coverage of this topic. %However, we refer the reader to the work of Assaad \emph{et al.} \cite{assaad2022survey} for a comprehensive description of the temporal extension of SCMs.

\subsection{SCM Architecture}

\begin{figure}[t]
    \centering
    \vspace{2.5mm}
    \includegraphics[width=0.8\linewidth]{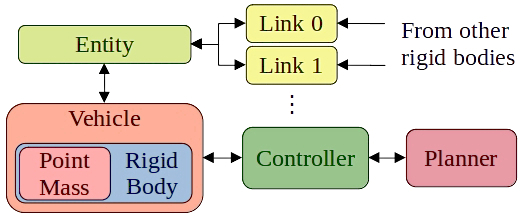}
    \caption{\footnotesize SCM architecture of the causal autonomous system for vehicles. 
    %A front-wheel drive vehicle representation inherits from a rigid body representation. This takes motor torque and steer as input from a controller that in turn takes action goals which are provided by a planner. Lastly, external forces are captured through an entity wrapper that connects the rigid body representation to other rigid body representations in the environment.
    }\vspace{-1mm}
    \label{fig:overall_architecture}
\end{figure}

We have built our methodology upon the SCM architecture presented in our previous work \cite{howard2024extending}. Here we proposed a series of extensions to the SCM formalism aimed at easing their integration with autonomous embodied systems. To demonstrate these extensions we carried out two case studies, with one providing a representation for AVs.

An overview of this representation is given in Fig. \ref{fig:overall_architecture}, with each node equating to an SCM module. While a full specification of SCMs is given in our previous work \cite{howard2024extending}, we provide an overview here for the benefit of the reader:
\subsubsection{Point Mass}
Represents a 2D point mass object \cite{blum2006mathematics} with the kinematic properties of position, linear velocity, and linear acceleration. The linear acceleration is derived from the mass of the object and the applied forces.
\subsubsection{Rigid Body}
Extends the \emph{Point Mass} SCM to capture a 2D rectangular rigid body with accompanying rotation, angular velocity, and angular acceleration. The angular acceleration is derived from the moment of inertia of the object and the applied torques.
\subsubsection{Vehicle}
Extends the \emph{Rigid Body} SCM to capture a front-wheel drive vehicle based upon a dynamic bicycle model \cite{guiggiani2018science}. This model takes motor torque and steering as inputs and calculates resulting forces and torques based upon the specifications of the vehicle.
\subsubsection{Entity}
Represents an interface between a \emph{Rigid Body} SCM and a shared environment, providing external forces and torques to the former from the latter. It captures both air resistance and collisions between rigid bodies. The air resistance is modelled entirely within the \emph{Entity} SCM, while the latter relies upon \emph{Link} SCMs to model calculations between pairs of \emph{Rigid Body} SCMs.
\subsubsection{Link}
Captures calculations between pairs of \emph{Rigid Body} SCMs, which are utilised by the \emph{Entity} SCM assigned to the primary \emph{Rigid Body} SCM. Namely, this SCM calculates collisions and distance headway between objects.
\subsubsection{Controller}
Calculates motor torque and steering to be passed to a \emph{Vehicle} SCM based upon an input action $a = (g_s, g_l)$, specified in terms of a speed goal $g_s$ and a lane goal $g_l$. A goal $g_x = (v_x, t_x)$ is specified in terms of a target value $v_x$ and a target time $t_x$. Based upon these parameters and the current time and vehicle state one can derive motor torque and steering via proportional-derivative control \cite{liptak2013process}.
\subsubsection{Planner}
Plans the current action $a$ to be passed to a \emph{Controller} SCM. To do so the planner produces a set of potential actions $\{\hat{a}_{0}, \hat{a}_{1}, ...\}$ and simulates outcomes $\{\hat{o}_{0}, \hat{o}_{1}, ...\}$ associated with these actions, using the generative properties of the SCMs. The planner then selects the best action based off reward function $r^\ast(\cdot)$ (see Sec. \ref{sec:learning_reward_parameter_profiles}).

%We base our work upon the architecture developed by Howard and Kunze \cite{howard2024extending}, an overview of which is shown in Fig. \ref{fig:overall_architecture}. Here each individual vehicular agent is represented by a dynamic bicycle \cite{guiggiani2018science} SCM extended from a rigid body SCM. These components are managed by controller SCMs that take a high-level action as input, this action in turn coming from a greedy planner SCM. Lastly the dynamics of each vehicular agent are tied together with a central collection of SCMs capturing environmental and inter-agent dynamic interactions. The result is a causal model fully capturing the dynamics, control and planning of vehicular agents, thus providing transparency and facilitating causal inference at multiple levels of granularity. For the purposes of this work, we interact with the SCM architecture by intervening upon the action variables of agents found in the planner layer, and by utilising the generative properties of the SCMs to simulate outcomes.

%===============================================================================

\section{Learning Reward Profiles} \label{sec:learning_reward_parameter_profiles}
We assume that the vehicular agents we consider represent intentional systems \cite{dennett1971intentional} and as such act in order to maximise some kind of conception of reward. Calculations for reward are derived from an outcome $o = (lt, f\!s, dh, e\!f, ad)$, which tracks the lane transitions ($lt \in \mathbb{Z}$), final speed ($f\!s \in \mathbb{R}$), distance headway ($dh \in \mathbb{R}$), and maximum environmental force magnitude ($e\!f \in \mathbb{R}$) of vehicle $A$, as well as whether the action in question had its goals accomplished ($ad \in \mathbb{B}$). 
%Given an outcome, 
We calculate reward as:
\begin{equation} \label{eq:reward_calc}
    r^\ast(o) = 
        \mathbf{r}(o)
        \mathbf{p}
\end{equation}
\begin{equation}
    \mathbf{r}(o) = \begin{bmatrix}
        r_{0}(o) & r_{1}(o) & r_{2}(o) & r_{3}(o) & r_{4}(o) & 1
        \end{bmatrix}
\end{equation}
where $\mathbf{p} = \begin{bmatrix}\gamma_0&\gamma_1&\gamma_2&\gamma_3&\gamma_4&\gamma_5\end{bmatrix}^\intercal$ is a reward profile comprised of weights. These weights are applied to a vector of reward metrics with a bias term $\mathbf{r}(\cdot)$ in order to calculate the overall reward associated with the outcome. The reward metrics capture a range of measures of utility that may be of varying levels of importance, and are defined as follows:
\begin{equation}
    r_0(o) = \sigma(-lt)
\end{equation}
\vspace{-2mm}
\begin{equation}
    r_1(o) = \min (\frac{dh}{\beta_{dh} \cdot f\!s},\ 1)
\end{equation}
\begin{equation}
    r_2(o) = e^{0.05 (f\!s - \beta_{f\!s})}
\end{equation}
\begin{equation}
    r_3(o) = e^{-0.05 f\!s}
\end{equation}
\begin{equation}
    r_4(o) = \delta_{e\!f \leq \beta_{e\!f}}
\end{equation}
Here $r_0(\cdot)$ captures reward associated with lane transitions using the sigmoid function $\sigma(\cdot)$. $r_1(\cdot)$ represents the desire for a vehicle to maintain safe distance from the vehicle in-front. Here a parameter $\beta_{dh} = 2\ s$ is used alongside the final speed to mirror the two-second rule \cite{rsa2012twosecond} often recommended by driving authorities. $r_2(\cdot)$ and $r_3(\cdot)$ reflect the goals of achieving a faster / slower final speed respectively, with $\beta_{f\!s}=31.3\,m/s$ acting as a speed-limit. Lastly, $r_4(\cdot)$ uses an altered Kronecker delta to give 1 if external forces imposed on the vehicle exceed $\beta_{e\!f}\,=\,1000\,N$ and 0 otherwise.

Since we are assuming that as an intentional system the agents are acting to maximise some concept of reward, we can try to infer $\mathbf{p}$ by considering the choice of action for a given agent. This bears some resemblance to established ideas within IRL \cite{arora2021survey}, although we are in only interested in the weightings at the moment a decision is made by the agent, rather than learning for use in an on-going policy. It is worth noting that while a linear weighting vector may be quite a simple representation, it has nonetheless found use in contemporary works \cite{gyevnar2024causal}. The primary logic is that linear weightings are inherently easy to for humans to interpret, an important trait for work aiming to provide explanations.
%Furthermore the novelty of this work is in the synthesis of this approach to reward profile learning combined with the causally grounded generation of explanations, rather than its use in isolation.

With this established, for a given action $a$, we consider a set of possible actions $\{\hat{a}_{0}, \hat{a}_{1}, ...\}$ that an agent could have taken at the time $a$ was executed. Using the generative properties of the SCM architecture, we simulate the outcomes associated with each action $\{\hat{o}_{0}, \hat{o}_{1}, ...\}$. We now pass each of these outcomes along with the observed outcome $o$ into a distance function:
\begin{equation} \label{eq:outcome_distance}
\begin{split}
                d(o, o^\prime) = \alpha_{o} (
                 \alpha_{lt} (lt - {lt}^\prime)^2 
                 + \alpha_{f\!s} (\frac{2(f\!s - {f\!s}^\prime)}{f\!s + {f\!s}^\prime})^2 \\
                 +\ \alpha_{dh} (\frac{dh}{f\!s} - \frac{{dh}^\prime}{{f\!s}^\prime})^2
                 + \alpha_{e\!f} (e\!f - {e\!f}^\prime)^2 \\
                 +\ \alpha_{ad} (ad - {ad}^\prime)^2
                 )^{\frac{1}{2}}
\end{split}
\end{equation}
%\begin{equation}
%    \omega_{lt}(lt, {lt}^\prime) = \alpha_{lt} (lt - {lt}^\prime)^2 
%\end{equation}
%\begin{equation}
%    \omega_{f\!s}(f\!s, {f\!s}^\prime) = \alpha_{f\!s} (\frac{2(f\!s - {f\!s}^\prime)}{f\!s + {f\!s}^\prime})^2 
%\end{equation}
where $\alpha_{o} = 0.1$, $\alpha_{lt} = 100$, $\alpha_{f\!s} = 1$, $\alpha_{dh} = 0.1$, $\alpha_{e\!f} = 0.01$, and $\alpha_{ad} = 100$ are scaling parameters designed to weight the importance of difference in each aspect of the outcome. Through this comparison we assign each hypothetical outcome an overall reward based upon the negative exponent of the distance function. Along with (\ref{eq:reward_calc}), we can formulate this as a linear regression task:
\begin{equation} \label{eq:linear_regression}
        \begin{bmatrix}
        \mathbf{r}(\hat{o}_0)\\
        \mathbf{r}(\hat{o}_1)\\
        \vdots\\
        \end{bmatrix}
        \mathbf{p}\ 
        =
        \begin{bmatrix}
            e^{-d(o,\hat{o}_0)} \\
            e^{-d(o,\hat{o}_1)} \\
            \vdots
        \end{bmatrix}
\end{equation}
From here we utilise the Householder rank-revealing QR decomposition with column pivoting approach implemented by Eigen \cite{gael2010eigen} to provide a solution for $\mathbf{p}$. The reward profile $\mathbf{p}$ should offer insight into the motivations behind the agent choosing $a$, given that the hypothetical actions with outcomes closest to $o$ would have been assigned the highest rewards on the right-hand side of (\ref{eq:linear_regression}). Importantly for this work, this enables us to more accurately reason how the agent's behaviour may have altered had circumstances differed at the time of decision making. This then allows the generation of causal explanations through counterfactual inference.

%===============================================================================

\section{Generating Causal Explanations of Agent Behavioural Interactions} \label{sec:causal_behaviour_explanation}

The goal of this work is to utilise the learnt reward profiles in order to better explain vehicular agent behavioural interactions. Hence we aim to establish explanations in the form of causal links between actions, where an action $a_C$ of agent $C$ was necessary for agent $A$ to select action $a_A$. 

The real-world vehicle data we consider is mostly given in terms of continuous variables in time series. Thus we utilise the approach described in our previous work \cite{howard2023simulation} to extract discrete time-action pairs, where the time part of the pair indicates the time $t_a$ at which the action was taken. 
%This approach defines a vehicular action $a = (\mathcal{g}_s, \mathcal{g}_l)$ as being comprised of a speed goal $\mathcal{g}_s = (v_s, t_s)$ and a lane goal $\mathcal{g}_l = (v_l, t_l)$. Each of these goals describe a target value (e.g. target speed, target lane id), and a time by which the vehicle aims to meet said target. 
Once we have extracted the actions for vehicles using the aforementioned approach, we can iterate over pairs of actions and test for the presence of causal necessity. Here we can additionally utilise the property of temporal precedence --- i.e. cause must come before effect in time --- to limit the number of pairs that need be considered.

In order to test for the presence of causal necessity we follow the approach depicted in Fig. \ref{fig:twin_world} to consider the decision-making process of $A$ via counterfactual inference. The first step in achieving this is to obtain a reward profile $\mathbf{p}_A$ for the time-action pair $(t_{a_A}, a_A)$ using the process detailed in Sec. \ref{sec:learning_reward_parameter_profiles}, allowing us to emulate the planning process of $A$ while utilising a similar conception of reward.

We now plan for $A$ at time $t_{a_A}$ under two worlds. The first is the observed world $\mathcal{W}$, while the second is a simulated version of the world $\mathcal{W}^{\neg C}$ in which $C$ never executed $a_C$. We derive $\mathcal{W}^{\neg C}$ by intervening upon the upon the input to the \emph{Controller} SCM such that the previous action of $C$ is maintained. The resulting distributions of the overall causal model are altered accordingly.

%We plan for $A$ under the worlds described by $\mathcal{W}$ and $\mathcal{W}^{\neg C}$ producing actions $\tilde{a}_A$ and $\tilde{a}_A^{\neg C}$ respectively. 
For each world \emph{Planner} SCM takes a range of possible actions $\{\hat{a}_{0}, \hat{a}_{1}, ...\}$ and simulates them by intervening upon the input action of the \emph{Controller} SCM. The SCM is then used to generate outcomes $\{\hat{o}_{0}, \hat{o}_{1}, ...\}$ for a predefined simulation horizon $\tau$. From here one can apply (\ref{eq:reward_calc}) to the outcomes in order to derive the best actions $\tilde{a}_A$ and $\tilde{a}_A^{\neg C}$ for worlds $\mathcal{W}$ and $\mathcal{W}^{\neg C}$ respectively.
%The planner here relies upon simulation of a set of potential actions through the generative properties of the SCM architecture; reward calculation of the resulting outcomes is performed with (\ref{eq:reward_calc}) and $\mathbf{p}_A$; followed finally by selection of the maximum reward action. However, the overall approach is planning agnostic, so long as each round is seeded the same across the two worlds for the purposes of planning reliant on random sampling.

We now pass $\tilde{a}_A$ and $\tilde{a}_A^{\neg C}$ to a distance function:
\begin{equation} \label{eq:action_distance}
\begin{split}
                d(a, a^\prime) = \alpha_{a} (
                (\frac{2(v_s\!-\!{v_s}^\prime)}{v_s\!+\!{v_s}^\prime})^2
                + (t_s\!-\!{t_s}^\prime)^2\\
                +\ \alpha_{v_l} (1\!-\!\delta_{v_l {v_l}^\prime})
                + (t_l\!-\!{t_l}^\prime)^2
                )^{\frac{1}{2}}
\end{split}
\end{equation}
where $\alpha_{a} = 0.1$ is a scaling parameter for the whole function and $\alpha_{v_l} = 10$ determines the distance to attribute to differing lanes, as indicated by the Kronecker delta $\delta_{v_l {v_l}^\prime}$. Provided the output of (\ref{eq:action_distance}) is greater than a predetermined threshold $\lambda_{a}$, we determine that the actions are sufficiently different, and thus $C$ executing $a_C$ caused $A$ to select $a_A$.

By determining that causality exists between two actions one can construct a causal graph with agent actions as the vertices and directional causal links between actions as edges that describe the causal influence that an agent taking one action had on an action taken by another agent. The resulting causal graph along with the reward profiles for the relevant agent actions describe how and why the scene unfolded as it did, both in terms of agent interactions and motivations.

%===============================================================================

\section{Experiments} \label{sec:experiments}

\subsection{Quantitative} \label{subsec:quantitative}

\paragraph{Datasets}
The dataset used in our evaluation is the highD dataset \cite{krajewski2018highd}. It consists of vehicle tracks extracted from labelling of overhead footage of German highways. Given a wide range of meta-data and the simple map structure of several lanes parallel to one-another, it allows for automatic extraction of scenarios with pre-established causal links. %The simplicity of the map is of key importance however, as more complex maps make it harder to definitively say whether two agents are interacting when determining which scenes to extract.

\paragraph{Scenario}
For the qualitative assessment of the proposed methodology and baselines, we consider a series of scenes in which two vehicular agents within the same lane are in close proximity to one another. The front vehicle (\emph{c0}) can affect the behaviour of the rear vehicle (\emph{c1}) by accelerating / decelerating, while the rear vehicle can affect the front vehicle through actions such as tailgating. Thus we consider these agents to possess causal adjacency, as they affect each other's behaviour. A further set of entirely independent vehicles (\emph{i0}, \emph{i1}, ...) are included in the scene in order to ensure the methods are precise in their detection of behavioural interactions between agents. This scenario is used because it is possible to automate the extraction of scenes matching its conditions based upon highD dataset metadata. Through this we extract 115 scenes for evaluation.
%Using these conditions we extract a total of $115$ scenes from the highD dataset for use in evaluation. %The majority of these scenes capture $5$ independent vehicles, which along with the $2$ convoy vehicles can mean up to $7$ vehicles in a scene. As such, for each scene the evaluated methods need to assess the validity of up to $21$ causal relationships between vehicles, with only the causal relationship between convoy vehicles being in fact valid.

\paragraph{Evaluation Metrics}
Receiver operating characteristic metrics such as precision, recall / true-positive rate (TPR), false-positive rate (FPR) and $\text{F}_1$ score. Here recall / TPR measures the proportion of true causal links identified as such, FPR measures the proportion of non-causal links falsely identified as causal links, and precision measures the proportion of links identified as causal that are true causal links. Lastly, $\text{F}_1$ score is the harmonic mean of precision and recall. 
%The means of calculating these from true-positive (TP), false-positive (FP), true-negative (TN) and false-negative (FN) counts is well documented. 
We derive these counts are calculated as follows:
%We can now formally define the means of calculating the true-positive $TP$, false-positive $FP$, true-negative $TN$, and false-negative $FN$ counts:
\begin{equation}
    (|TP|,|FP|,|FN|) = (|\hat{G} \cap G|, |\hat{G} \setminus G|, |G \setminus \hat{G}|)
\end{equation}
%\begin{equation}
%    |FP| = |\hat{G} \setminus G|
%\end{equation}
%\begin{equation}
%    |FN| = |G \setminus \hat{G}|
%\end{equation}
\begin{equation}
    |TN| = \frac{|\mathcal{V}| (|\mathcal{V}| - 1)}{2} - (|TP| + |FP| + |FN|)
\end{equation}
where $G$ and $\hat{G}$ are the ground truth and predicted causal adjacency graphs respectively, and $\mathcal{V}$ is the set of vehicular agents common to each adjacency graph. The causal adjacency graph $\hat{G}$ is built by creating an edge between two agents if there is a cause-effect relation between any of their actions. Meanwhile the adjacency edges of $G$ consists solely of the non-directional edge between agents $\emph{c0}$ and $\emph{c1}$, corresponding to the scenario described above.

\paragraph{Implementation \& Baselines}
We provide access to our code, parameters, and output data / videos online\,\footnote{\href{https://github.com/cognitive-robots/gce_vbai_lrp_paper_resources}{\url{https://github.com/cognitive-robots/gce_vbai_lrp_paper_resources}}}. The proposed method is evaluated against the three variants of our previous work, SimCARS-v1 \cite{howard2023simulation}; Multi-Variate Granger causality \cite{geweke1982measurement}; TiMINo \cite{peters2013causal}; and DYNOTEARS \cite{pamfil2020dynotears}. 
These represent our previous iteration of work along with the three best performing temporal causal discovery methods for agent behavioural interactions based upon a benchmark \cite{howard2023evaluating}.

\paragraph{Parameters}
Non-SimCARS methods were tested for thresholds of $0.001$, $0.005$, $0.01$, $0.03$, $0.05$, and $0.1$. The SimCARS-v1 reward-based and hybrid variants were tested for thresholds of $\{ 0.1\alpha | \alpha \in [1..10] \}$. Lastly SimCARS-v2 was tested for thresholds of $\{ 0.1\alpha | \alpha \in [0..10] \}$.

\paragraph{Results}

\begin{figure}[t]
    \centering
\begin{subfigure}[t]{\linewidth}
    \centering
    \vspace{2.5mm}
    \includegraphics[width=0.925\linewidth]{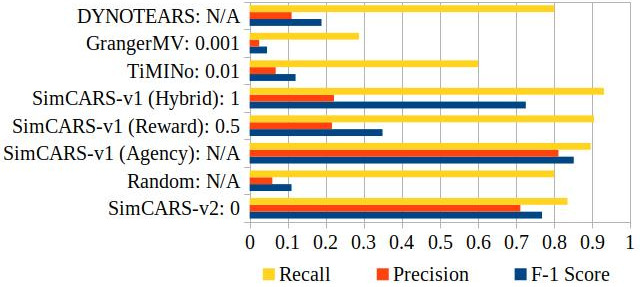}
    \caption{\footnotesize The precision, recall and $\text{F}_1$ score of evaluated methods. Numbers suffixed to the methods show the threshold giving the max $\text{F}_1$ score depicted.}
    \label{fig:bar_graph}
\end{subfigure}\\[1mm]
\begin{subfigure}[t]{\linewidth}
    \centering
    \includegraphics[width=0.85\linewidth]{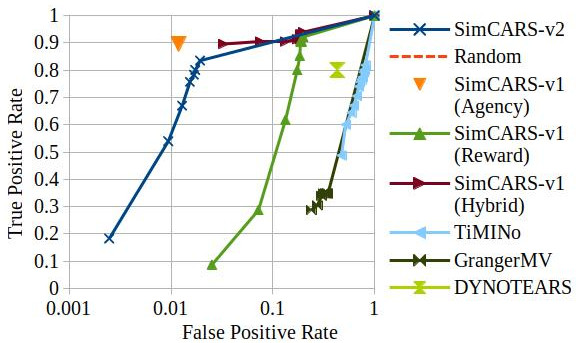}
    \caption{\footnotesize Receiver operating characteristic (ROC) curve. Single points represent methods without a threshold. The x-axis is logarithmic for the sake of clarity.}
    \label{fig:roc_curve}
\end{subfigure}

\caption{\footnotesize Quantitative Results}
\label{fig:quantitative}
\end{figure}

In Fig. \ref{fig:quantitative} the quantitative experiment results are shown. The proposed method (SimCARSv2) is highly competitive against existing methods, with only the agency-based variant of SimCARSv1 scoring higher than the proposed method in $\text{F}_1$ score (see Fig. \ref{fig:bar_graph}). Importantly the proposed method demonstrates a massive improvement over the SimCARSv1 reward-based variant in terms of precision, which is the most similar method --- both rely upon a weighting of reward metrics. This demonstrates that the linear regression approach to learning reward metric weightings is effective in capturing the priorities of autonomous agents.

Overall the main factor that limits the performance of SimCARSv2 is it's comparatively low sensitivity, given that it gives a maximal $\text{F}_1$ score and recall of $0.768$ and $0.835$ respectively for a threshold $\lambda_a > 0$.
%, i.e. any difference in action is sufficient for a causal link to test positive. 
A positive of this is that for higher thresholds SimCARSv2 is more precise than even the agency-variant of SimCARSv1 as indicated by the ROC curve (see Fig. \ref{fig:roc_curve}), which could make it useful in situations where precision is of greater importance. However, given that the method is largely precise across thresholds, the practical choice is to select a threshold of zero, maximising the recall at relatively little expense to the precision.

A potential cause of the comparatively lower sensitivity could be the reward metrics comprising the overall reward function. If the reward metrics present are not expressive enough to capture the motivations of the agent in question, then other metrics may end up being utilised as proxies when estimating the reward profile. If one then attempts to use this reward profile during counterfactual inference the agent cognition might significant deviate from the original agent, in turn leading it to overlook certain causal relationships. Of course, the dependence of the method performance on a suitable set of reward metrics is indeed a limitation, as this may be hard to infer during system design.

Another limiting factor is a lack of information with which to refine simulation / agent parameters. While SimCARSv2 offers an enhanced level of accuracy over SimCARSv1 in terms of it's dynamics modelling, the ability to exploit this is limited by the type of meta-data available in the highD dataset. On a real-world deployment of SimCARSv2 one could configure the parameters of the vehicles based upon vehicle specifications, rather than relying upon rough approximations. Furthermore the utilisation of an SCM architecture allows the use of distributions rather than fixed values for model inputs, something not possible with SimCARSv1.
%Given that the agency-based variant of SimCARS is effectively based upon a subset of the metrics considered by the proposed approach it should hypothetically be possible to match or even surpass the performance of this baseline through further parameter tuning. This does however indicate that a potential weakness of the proposed approach is the number of parameters, and future work to simplify the means of outcome comparison could help alleviate the situation.

\subsection{Qualitative} \label{subsec:qualitative}

\paragraph{Datasets}
In addition to the highD dataset \cite{krajewski2018highd} introduced previously, we also utilise the exiD \cite{moers2022exid} and inD \cite{bock2020ind} datasets to provide a wider range of scenarios. These are similarly captured and formatted to the highD dataset, but instead of focusing on typical highway stretches, they consider on / off ramps and intersections respectively.

\paragraph{Results}

\begin{figure*}[t]
    \centering

    \vspace{2.5mm}
    \begin{subfigure}{0.32\linewidth}
    \centering
    \includegraphics[width=\textwidth]{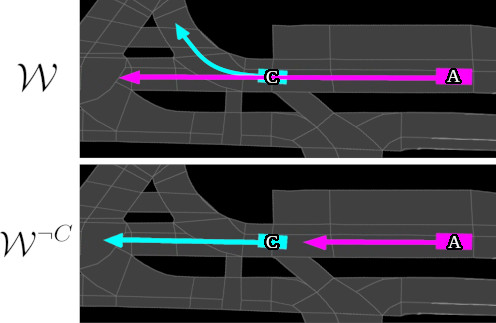}
    \caption{\footnotesize inD - Accelerating / Braking}
    \label{fig:ind_qualitative}
    \end{subfigure}
    \quad
    \begin{subfigure}{0.55\linewidth}
    \centering
    \includegraphics[width=\textwidth]{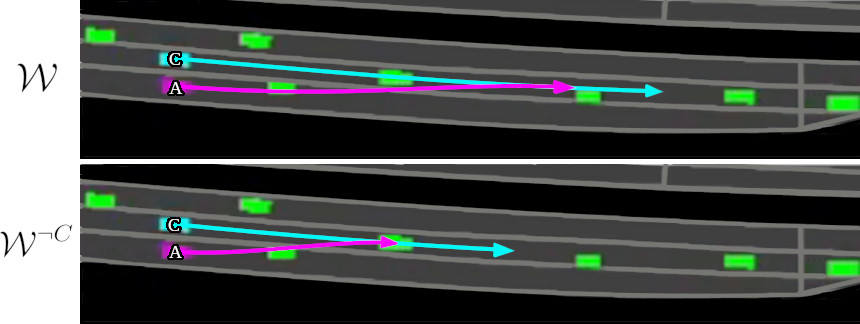}
    \caption{\footnotesize exiD - Merging}
    \label{fig:exid_qualitative}
    \end{subfigure}\\[0.2cm]
    
    \begin{subfigure}{0.625\linewidth}
    \centering
    \includegraphics[width=\textwidth]{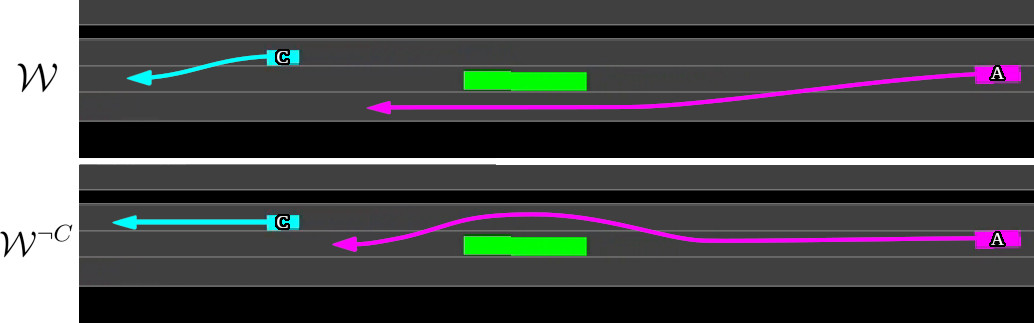}
    \caption{\footnotesize highD - Overtaking}
    \label{fig:highd_qualitative}
    \end{subfigure}
    \quad
    \begin{subfigure}{0.265\linewidth}
    \centering
    \includegraphics[width=\textwidth]{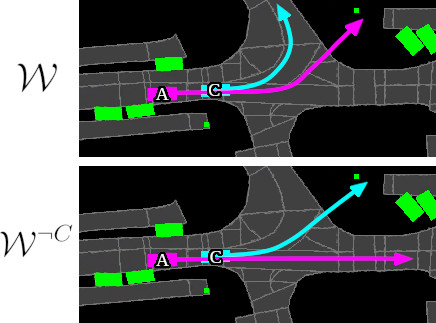}
    \caption{\footnotesize inD 2 - Failure Case}
    \label{fig:failure_qualitative}
    \end{subfigure}
    
    \caption{\footnotesize Illustration of twin-world analysis of driving scenes. $\mathcal{W}$ denotes the planned behaviour under the original world state at the time the affected action $a_A$ was taken. Meanwhile $\mathcal{W}^{\neg C}$ denotes the planned behaviour under the counterfactual world state in which the causing action $a_C$ was not taken, at the same time as before. The magenta vehicle indicates the affected agent, the cyan vehicle the causing agent, and green vehicles the background agents.
    %The circular nodes indicate the spatial points at which behaviour diverges between the counterfactual twin-worlds and the observed world. For the causing agent, there is just the observed path (solid line) and the simulated path without the causing action (dotted line). For the affected agent, there is the observed path (solid line), the simulated path with the causing action (single dot dashed line), and the simulated path without the causing action (double dot dashed line).
    \vspace{-2mm}}
    \label{fig:qualitative}
\end{figure*}

\begin{figure}
\vspace{2mm}
    \centering
    \includegraphics[width=\linewidth]{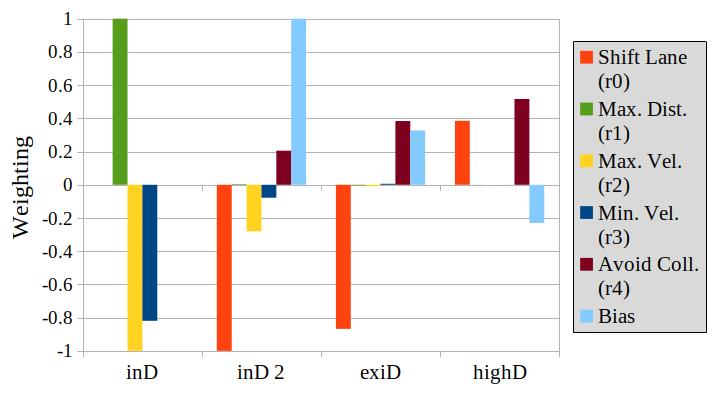}
    \caption{\footnotesize Reward Profiles for Each Scenario}
    \label{fig:profiles_qualitative}
\vspace{3mm}
\end{figure}

Here we select three scenes in particular for examination with the goal of exploring a variety of interaction types. These are depicted in Fig. \ref{fig:qualitative}.

The first of these presents an intersection where agent $C$ turns right while agent $A$ continues on ahead, accelerating as it does (see Fig. \ref{fig:ind_qualitative}). 
%Whether the causing agent turns right or not, the affected agent will continue straight ahead. 
However, the SimCARS-v2 suggests that if agent $C$ continues straight agent $A$ will instead slow down.
%--- no longer having a clear path to accelerate ahead. 
The reward profile of this scenario (see Fig. \ref{fig:profiles_qualitative}) demonstrates that agent $A$ both wishes to maintain a particular speed --- determined via $r_2$ and $r_3$ --- and maximise its distance headway --- via $r_1$, which further justifies the behavioural seen during counterfactual inference. 
%This justifies the behaviour seen in both twin-worlds as agent $A$ will aim to accelerate to its desired speed, but only if it does not decrease its distance headway.

The second scene has agent $A$ merging from an on-ramp into the lane occupied by agent $C$ (see Fig. \ref{fig:exid_qualitative}) with agent $C$ accelerating just before the merge. 
%However, in this scene the causing agent accelerates moments before the merge taking place. 
SimCARS-v2 suggests that without this acceleration taking place, agent $A$ would have to slow down before merging, or else risk a collision. %Hence a behavioural interaction is present between the agents. 
The reward profile (see Fig. \ref{fig:profiles_qualitative}) indicates agent $A$ wishes to shift lane --- via $r_0$ --- while avoiding a collision --- through $r_4$, in keeping with the twin-world simulation behaviour. 
%Again this justifies the behaviour seen in each of the twin-worlds as agent $A$ aims to shift lane in each case, but must brake for case $\mathcal{W}^{\neg C}$ in order to not collide with agent $C$.

The third scene depicts agent $A$ overtaking another vehicle (see Fig. \ref{fig:highd_qualitative}) shifting left to do so. SimCARS-v2 infers that this course of behaviour was the result of a lane change by agent $C$. Otherwise agent $A$ would have found it preferable to shift right before moving to the lane agent $C$ would end up in. The reward profile (see Fig. \ref{fig:profiles_qualitative}) is similar to the previous except with the lane bias flipped, indicating that agent $A$ deciding to shift left over right was primarily driven by a desire to avoid collision with agent $C$ following its lane shift.
%This again makes sense, given that agent $A$ in case $\mathcal{W}$ would avoid overtaking to prevent collision with agent $C$, preferring to shift into a lane absent of vehicles. Meanwhile agent $A$ in case $\mathcal{W}^{\neg C}$ can carry out an overtaking maneuver without risk of collision with agent $C$, given that agent $C$ remains in its original lane for this case.

\vspace{3mm}
The final scene shows a failure case for SimCARS-v2 (see Fig. \ref{fig:failure_qualitative}). Here the agents in question likely do have a behavioural interaction, which the system identifies. However, the planned actions via which it determines this are nonsensical, with an agent veering off the road in each case. This is likely the result of the default behaviour at the branching point of a lane being undefined, and the reward metrics not taking into account lane following, issues that should be addressed in future work.
%This is a combination of two issues. The first being that the behaviour of agents when they reach the end of a lane segment is undefined, e.g. if we intervene to prevent an agent from turning left at an intersection, then does it turn right or go straight? The second issue is that because whether or not the vehicle stays on the road is not part of the outcome or reward metrics, the agent will happily drive the vehicle off of the road provided it maximises its reward. This highlights the importance of defining a diverse and comprehensive reward model.
\vfill
%===============================================================================

\section{Discussion \& Future Work}
%Despite SimCARSv2 largely being effective, it does have its limitations. While the issue of undefined behaviour following lane segments is raised, this could be solved on a real-world system where destination data is present. The greater challenge is in improving the expressiveness of the reward profiling of agent decisions.%The bigger issue is the lack of expressiveness in the outcome / reward model, which effectively permits any behaviour so long as it maximises reward. We also made assumptions about control and planning, and where the dividing line lies between them.

Based upon the types of failure case discussed in the last section, it seems sensible to consider alternate means of modelling and approximating agent action motivations. %A challenge is that any variable we wish to consider as part of a reward model must be captured by the SCM architecture. %e.g. distance headway had to be explicitly implemented within the SCM to be used in the reward model. 
The challenge of deciding which information to include in such a model is ultimately a design decision. However, one could potentially train some form of data-driven model (e.g. a neural net) that could be used to sanity check the current weighted reward model. One may want to avoid replacing the weighted reward model entirely as the primary reason for selecting said model was due to its inherent interpretability.

%\subsection{Future Work \& Conclusion}
%In terms of next steps, there are a number of potential directions in which the proposed explanation methodology could be improved. Expanding the coverage of reward metrics and incorporating spatial information into simulation outcomes could both offer improvements. A failure to offer a diverse range of reward metrics which can result in the metrics that are present being used as inadequate proxies. And while simulation outcomes do consider lane transitions, it is still possible for outcomes which have notably different final positions being considered similar nonetheless.
%were both identified as potential extensions in Sec. \ref{subsec:analysis}, although beyond this an ablation study to optimise the parameters associated with such an approach could yield improved efficacy.

One could also explore integrating behavioural interaction causal modelling into a RL loop, potentially allowing for greater efficiency and socially-awareness when operating around humans. This would differentiate itself from existing work merging causal reasoning and RL \cite{bareinboim2015bandits,gasse2022causal,he2023causal} by its behavioural interaction focus over typical egocentric perspectives. This is particularly relevant for domains involving a great deal of interaction with humans.

In this work we have demonstrated how information regarding the motivations of a vehicular agent can be incorporated into twin-world counterfactual inference in order to detect causal behavioural interactions. We also show how a reward profile representing the instantaneous motivations of a vehicular agent can be approximated via simulation and linear regression. We have demonstrated via quantitative experiments that this approach is competitive against previous work, and significantly improves over the next-best reward-based model. Furthermore we illustrate several scenarios which show the capabilities and limitations of the approach via qualitative experiments. Overall this work represents another step towards the responsible development of AVs, adhering to tenants such as explainability and transparency.

%===============================================================================

\addtolength{\textheight}{0cm}   % This command serves to balance the column lengths

%%%%%%%%%%%%%%%%%%%%%%%%%%%%%%%%%%%%%%%%%%%%%%%%%%%%%%%%%%%%%%%%%%%%%%%%%%%%%%%%

%%%%%%%%%%%%%%%%%%%%%%%%%%%%%%%%%%%%%%%%%%%%%%%%%%%%%%%%%%%%%%%%%%%%%%%%%%%%%%%%

%%%%%%%%%%%%%%%%%%%%%%%%%%%%%%%%%%%%%%%%%%%%%%%%%%%%%%%%%%%%%%%%%%%%%%%%%%%%%%%%
%\section*{APPENDIX}


%\section*{ACKNOWLEDGMENT}

%The preferred spelling of the word ÒacknowledgmentÓ in America is without an ÒeÓ after the ÒgÓ. Avoid the stilted expression, ÒOne of us (R. B. G.) thanks . . .Ó  Instead, try ÒR. B. G. thanksÓ. Put sponsor acknowledgments in the unnumbered footnote on the first page.

%%%%%%%%%%%%%%%%%%%%%%%%%%%%%%%%%%%%%%%%%%%%%%%%%%%%%%%%%%%%%%%%%%%%%%%%%%%%%%%%

\bibliographystyle{IEEEtran.bst}
\bibliography{references.bib}

\end{document}